\def\year{2022}\relax
\documentclass[letterpaper]{article} 
\usepackage{aaai22}  
\usepackage{times}  
\usepackage{helvet}  
\usepackage{courier}  
\usepackage[hyphens]{url}  
\usepackage{graphicx} 
\usepackage{natbib}  
\usepackage{caption} 
\DeclareCaptionStyle{ruled}{labelfont=normalfont,labelsep=colon,strut=off} 
\usepackage{algorithm}
\usepackage{algorithmic}

\usepackage{xcolor}

\usepackage{newfloat}
\usepackage{listings}
\usepackage{amsmath}
\floatstyle{ruled}
\newfloat{listing}{tb}{lst}{}
\floatname{listing}{Listing}
\nocopyright
\title{TOD-DA: Towards Boosting the Robustness of Task-oriented \\ Dialogue Modeling on Spoken Conversations}
\author{
    Xin Tian\equalcontrib,
    Xinxian Huang\equalcontrib,
    Dongfeng He\equalcontrib,
    Yingzhan Lin,
    Siqi Bao,
    Huang He,
    Liankai Huang,
    Qiang Ju,
    Xiyuan Zhang,
    Jian Xie,
    Shuqi Sun,
    Fan Wang,
    Hua Wu,
    Haifeng Wang
}
\affiliations{
    Baidu Inc., China \\

    \{tianxin06, huangxinxian01, hedongfeng\}@baidu.com
}

\begin{document}

\maketitle

\begin{abstract}
Task-oriented dialogue systems have been plagued by the difficulties of obtaining large-scale and high-quality annotated conversations. Furthermore, most of the publicly available datasets only include written conversations, which are insufficient to reflect actual human behaviors in practical spoken dialogue systems. In this paper, we propose Task-oriented Dialogue Data Augmentation (TOD-DA), a novel model-agnostic data augmentation paradigm to boost the robustness of task-oriented dialogue modeling on spoken conversations. The TOD-DA consists of two modules: 1) \textit{Dialogue Enrichment} to expand training data on task-oriented conversations for easing data sparsity and 2) \textit{Spoken Conversation Simulator} to imitate oral style expressions and speech recognition errors in diverse granularities for bridging the gap between written and spoken conversations. With such designs, our approach ranked first in both tasks of DSTC10 Track2, a benchmark for task-oriented dialogue modeling on spoken conversations, demonstrating the superiority and effectiveness of our proposed TOD-DA.
\end{abstract}

\section{Introduction}

Task-oriented dialogue systems have been widely used in our daily life, especially in the voice assistants like Siri, Alexa, and Xiaodu. Nevertheless, these speech-based dialogue systems face several challenges. 
Firstly, given the expensive cost of conversation collection and annotation, it is difficult to obtain large-scale and high-quality training data for dialogue systems.
Secondly, most of the publicly available task-oriented dialogue datasets include only written conversations, which can not adequately reflect actual human behaviors in practical spoken dialogue systems. Involving extra noises from disfluency, grammatical mistakes and speech recognition errors, spoken conversations impose more challenges on the robustness of the dialogue system.

To address the problem of data insufficiency, some attempts have been made to expand the conversation dataset based on the original ones by rewriting the dialogues on word-level~\citep{wei-zou-2019-eda, liu-etal-2021-robustness} or sentence-level~\citep{iyyer2018adversarial, gao2020paraphrase,shleifer2019low}. However, these methods are confined by the dialogue structure of original datasets, and the expanded conversations lack diversity and flexibility on contents of dialogue-level. To bridge the gap between written and spoken modalities, previous methods try to imitate oral conversation style~\citep {liu-etal-2021-robustness} or simulate speech recognition  errors~\citep{gopalakrishnan2020neural, wang2020data}.
These works mainly focus on simulations on the word or sentence level, while the essential phoneme level is relatively under-explored.

In this paper, we propose Task-oriented Dialogue Data Augmentation (TOD-DA), a model-agnostic data augmentation paradigm for task-oriented dialogue on spoken conversations. The TOD-DA consists of two modules: Dialogue Enrichment (DE) and Spoken Conversation Simulator (SCS). Dialogue Enrichment aims to obtain a large amount of task-oriented conversations with annotations. Especially, the pattern based augmentation strategy in DE is not confined to the fixed dialogue structure of the existing datasets and is able to create diverse dialogues in a more flexible manner. Spoken Conversation Simulator tries to capture spoken conversation characteristics, which simulates oral style expressions and speech recognition errors in diverse granularities. With these two modules, the TOD-DA is able to boost the robustness of task-oriented systems on spoken conversations.

To benchmark the robustness of task-oriented dialogue modeling on spoken conversations, DSTC10 Track2~\citep{kim2021how} introduced two tasks: 1) \textit{multi-domain dialogue state tracking}, and 2) \textit{knowledge-grounded dialogue modeling}. In this track, without any training data, only a small amount of validation data was available, including the ASR outputs of the spoken conversations and the database. 
With the proposed TOD-DA, we ranked first place in both tasks of this track. The augmented dataset by TOD-DA has been released at GitHub, hoping to facilitate the future work of task-oriented dialogue modeling on spoken conversations.~\footnotemark[1]
\footnotetext[1]{\url{https://github.com/PaddlePaddle/Knover/tree/develop/projects/DSTC10-Track2}}

\section{Task-oriented Dialogue Data Augmentation}

\begin{figure}[ht]
	\centering
	\includegraphics[width=\columnwidth]{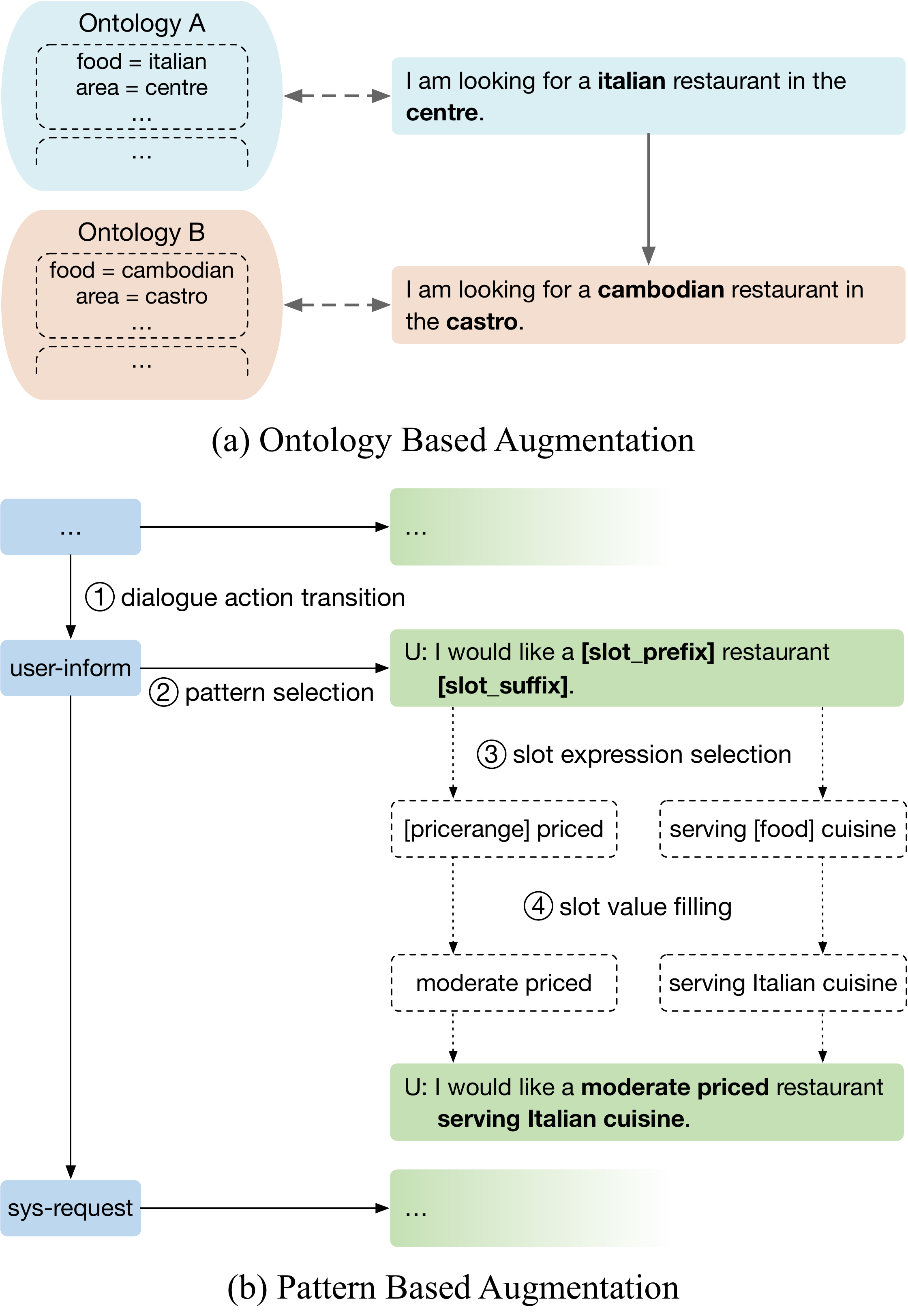}
	\caption{Illustrations of Dialogue Enrichment}
	\label{fig:de}
\end{figure}
In this paper, we propose Task-oriented Dialogue Data Augmentation (TOD-DA), a model-agnostic data augmentation paradigm for task-oriented dialogue systems on spoken conversations. Specifically, the TOD-DA consists of two modules: Dialogue Enrichment (DE) to expand conversation datasets, and Spoken Conversation Simulator (SCS) to bridge the gap between spoken and written conversation modalities.

\subsection{Dialogue Enrichment}

In dialogue enrichment, two strategies are introduced to obtain a large amount of task-oriented conversations: ontology based augmentation and pattern based augmentation.

\subsubsection{Ontology Based Augmentation}

Despite the insufficient training data in task-oriented dialogue systems, there exist some high-quality publicly available datasets, such as Frames~\citep{el-asri-etal-2017-frames}, WOZ~\citep{wen-etal-2017-network}, and MultiWOZ~\citep{budzianowski-etal-2018-multiwoz}. As such, in this paper, we propose to expand task-oriented conversations by fusing the existing datasets and the other ontology. Specifically, we will replace each slot value in the conversation with another value from the other ontology. As shown in Figure~\ref{fig:de}, the \textit{food} and \textit{area} in the original utterance are replaced by new values in \textit{Ontology B}. This is an efficient data enrichment strategy without changing the structure of the original high-quality conversations.

\subsubsection{Pattern Based Augmentation}

Although the above strategy ensures the correctness and fluency of the augmented conversations, it is confined to the fixed dialogue structure of the existing datasets.
Considering that training with less diverse datasets might be vulnerable to overfitting, we further propose a pattern based augmentation strategy, aiming to create diverse dialogues in a more flexible manner.

We decompose dialogues of public datasets into patterns according to the annotated dialogue actions such as \textit{inform}, \textit{request}, \textit{book}, etc. A pattern is typically a sentence with unfilled slots. For example, ``I would like a \texttt{[slot\_prefix]} restaurant \texttt{[slot\_suffix]}.'' is a pattern of \textit{user-inform} dialogue action. The slot in the pattern can be categorized into two types, namely prefix and suffix. The slot prefix is an adjective used before an entity like ``moderate priced'', whereas the slot suffix is a phrase placed after an entity like ``serving Italian cuisine''. The expression of each slot can be gathered by mining public datasets and exploring search engines. Furthermore, slots in the pattern are designed to be scalable, which means that a slot can be extended into multiple ones.

The procedure of pattern based augmentation can be divided into four steps: dialogue action transition, pattern selection, slot expression selection, and slot value filling. We regard the first step as a Markov chain, meaning that the selection of the next dialogue action will be restricted by the previous dialogue action according to a transition matrix. The transition probabilities are precomputed by normalizing the co-occurrence frequency of dialogue actions. In the first step, we employ a random walk based on the transition matrix to acquire dialogue actions. Secondly, we randomly sample a pattern for the selected dialogue action from its candidate pattern pool. Finally, we randomly choose an expression for each slot and fill the slot with values sampled from the database. At the same time, the selected slots and values are accumulated to the dialogue state annotation of the current turn.

This strategy is not constrained by the dialogue structure of the public datasets, improving the diversity of generated data. Theoretically, we can produce infinite task-oriented conversations, effectively alleviating the lack of training data.

\begin{table}[ht]
	\centering
	\includegraphics[width=\columnwidth]{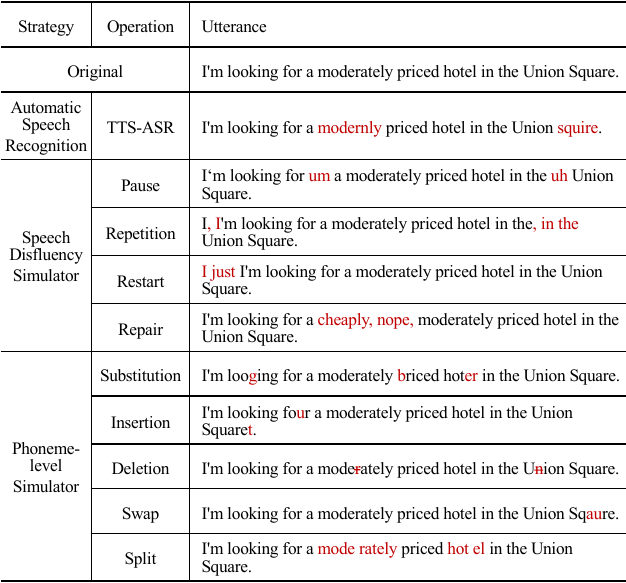}
	\caption{Examples of Spoken Conversation Simulator.}
	\label{tab:simulator-disturbance}
\end{table}

\subsection{Spoken Conversation Simulator}

Distinct from the written conversation, the spoken conversation tends to exhibit free expression styles. Besides, automatic speech recognition (ASR) in spoken conversations incurs a variety of errors. 
To bridge the gap between the written and spoken conversations, we propose Spoken Conversation Simulator (SCS), which is able to capture spoken conversation characteristics from multiple aspects. 
The SCS module consists of the three strategies: Automatic Speech Recognition, Speech Disfluency Simulator, and Phoneme-level Simulator.

\subsubsection{Automatic Speech Recognition}

We utilize a TTS-ASR pipeline to obtain noisy conversations given the written conversations. Firstly, the original text is transferred to audio with various accents by gTTS~\footnotemark[2], 
\footnotetext[2]{\url{https://pypi.org/project/gTTS/}}
a tool with Google Translate's text-to-speech API. Then, the audio is converted into text by wav2vec~\citep{NEURIPS2020_92d1e1eb} pre-trained on 960 hours of Librispeech~\citep{panayotov-chen-2015-librispeech}. As shown in Table~\ref{tab:simulator-disturbance}, some noises are inserted into utterances through this way, as a result of the similar word or phrase pronunciation.

\subsubsection{Speech Disfluency Simulator}

A prominent difference between written and spoken conversation lies in fluency. Similar to previous works~\citep{Wang_Che_Liu_Qin_Liu_Wang_2020, liu-etal-2021-robustness}, to simulate the disfluencies occurred in spoken conversation, we perform the following four operations:

\begin{itemize}
    \item \textbf{Pause} simulates speech pauses during communication by randomly inserting some filler words, such as ``uh'', ``um'', ``you know'', and so on.
    \item \textbf{Repetition} imitates the common behavior of repeating the previous words or phrases.
    \item \textbf{Restart} adds some common prefixes to the beginning of a sentence including ``I mean'', ``I just'', ``And'', etc.
    \item \textbf{Repair} mimics a slip of the tongue or intent mutation, which first provides a wrong slot value and then corrects it.
\end{itemize}

\subsubsection{Phoneme-level Simulator}

To further simulate ASR errors caused by similar pronunciation, we design phoneme-level simulator including the following five operations:

\begin{itemize}
    \item \textbf{Substitution} replaces letters with similar pronunciation in a word, for instance, ``b'' and ``p''.
    \item \textbf{Insertion} puts a random letter anywhere in the sentence.
    \item \textbf{Deletion} removes a letter from a word stochastically.
    \item \textbf{Swap} exchanges two adjacent vowels in the word randomly.
    \item \textbf{Split} divides words into several parts according to the length of words. This is motivated by the phenomenon that the ASR may divide a word into multiple parts, possibly due to the pauses within the word in the speech. For example, \textit{restaurant} becomes \textit{restau rant}, and \textit{moderately} becomes \textit{mode rately}.
\end{itemize}

\section{Task-oriented Dialogue Modeling}

In DSTC10, there is one track of task-oriented dialogue modeling on spoken conversations, which includes two tasks of multi-domain dialogue state tracking and knowledge-grounded dialogue modeling. We took part in this track to validate the effectiveness of the proposed TOD-DA.

\subsection{Multi-domain Dialogue State Tracking}

Dialogue state tracking (DST) is a crucial task in task-oriented dialogue systems, which extracts users’ goals at each turn of the conversation and represents them in the form of a set of (slot, value) pairs. A dialogue with $T$ turns can be represented as $D=\{D_1,D_2,\cdots,D_T\}$, where $D_t$ is the dialogue at turn $t$ consisting of system response and user utterance. Similarly, the dialogue state at turn $t$ is denoted as $B_t$ concatenated by a series of slots and values.

In this task, we rely on transformer-based dialogue pre-training models to generate dialogue state~\citep{vaswani2017attention, bao2020plato}. The most straightforward way is to take the whole dialogue history as input and generate the dialogue state as output. However, it is not efficient enough as they need to predict the dialogue state from scratch based on the dialogue history. On the merits of utilizing the previous dialogue state as a compact representation of the previous dialogue history, it is more effective and reasonable to use the previous dialogue state. In addition, we use the dialogue history within a window of size $n$ to assist the model in better capturing contextual information. Our dialogue state generation process can be written as:
\begin{equation}
    X_t=[D_{t-n},\cdots,D_t;B_{t-1}]
\end{equation}
\begin{equation}
    B_t=\text{Transformer}(X_t)
\end{equation}
where the input sequence $X_t$ is concatenated with recent dialogue history and the previous dialogue state.

\subsection{Knowledge-grounded Dialogue Modeling}

In task-oriented dialogue systems, some user requests are not covered by pre-defined domain APIs~\citep{kim-etal-2020-beyond}. By grounding on external knowledge sources, the coverage of task-oriented dialogue systems can be expanded. This task aims to model task-oriented conversation with unstructured knowledge access, and it includes three subtasks: \textit{Knowledge-seeking Turn Detection}, \textit{Knowledge Selection} and \textit{Knowledge-grounded Response Generation}. 

\subsubsection{Knowledge-seeking Turn Detection}

The system first should determine whether to trigger external knowledge access or not. A dialogue with $T$ turns can be represented as $D=\{D_1,D_2,\cdots,D_T\}$, where $D_t$ is the dialogue at turn $t$ consisting of system response and user utterance. This knowledge-seeking turn detection can be treated as a binary classification task, given the dialogue context $\{D_1,\cdots,D_t\}$ as input. The training objective is to minimize the binary cross-entropy loss:
\begin{equation}
\begin{split}
\mathcal{L}&_{\rm detection} = -y_t\log{p_{\rm detection}(\hat{y_t} = 1 | D_1,\cdots,D_t)}\\
&-(1-y_t)\log{(1 - p_{\rm detection}(\hat{y_t} = 1 | D_1,\cdots,D_t))}
\raisetag{1.65\baselineskip}
\end{split}
\end{equation}
where $y_t$ denotes the ground-truth label to trigger external knowledge access or not at turn $t$.

\subsubsection{Negatives Enhanced Knowledge Selection}

Once determined to trigger external knowledge access, the system needs to select the appropriate knowledge snippet by estimating the relevance between the given dialogue context and each knowledge snippet in the whole knowledge base.

Following the previous work \citep{he2021learning}, multi-scale negatives are utilized to boost the relevance estimation. Since models might mix up semantically similar knowledge snippets, we further consider semantic similarity during negative sampling. 
In summary, for each positive sample, we drew its negative samples from five scales: 
(1) Random: one knowledge snippet is randomly selected from the whole knowledge base; (2) In-Domain: one knowledge snippet is randomly selected from those within the same domain as the positive sample; (3) In-Entity: one knowledge snippet is randomly selected from those belonging to the same entity as the positive sample; (4) Cross-Entity: one knowledge snippet is randomly selected from those belonging to the aforementioned entity in the dialogue context; (5) In-Semantics: one knowledge snippet is randomly selected from those having high semantic similarity with the positive sample.

Formally, the external knowledge base is denoted as $K = \{k_1,k_2,\cdots,k_n\}$, where $k_i$ represents the $i$-th knowledge snippet. We estimate the following probability $p_{\rm selection}(l_{k_i} = 1 | D_1,\cdots,D_t, k_i)$, where $l_{k_i}$ stands for the label to select $k_i$ or not given the dialogue context. During training, the objective is to minimize the following loss:
\begin{equation}
\begin{split}
\mathcal{L}&_{\rm selection} = -\log{p_{\rm selection}(l_{k_i} = 1 | D_1,\cdots,D_t, k_i)}\\
&~~~~~~-\sum_{j}\log{p_{\rm selection}(l_{k_{i,j}^-} = 0 | D_1,\cdots,D_t, k_{i,j}^-)} 
\raisetag{1.65\baselineskip}
\end{split}
\end{equation}
where $k_{i,j}^-$ denotes the $j$-th negative sample of the positive sample ${k_i}$. During inference, the optimal knowledge snippet can be selected as:
\begin{equation}
k^* = \operatorname*{argmax}_{k_i\in K}p_{\rm selection}(l_{k_i} = 1 | D_1,\cdots,D_t, k_i)
\label{eq:obj_sel}
\end{equation}

\subsubsection{Knowledge-grounded Response Generation}

Finally, the system needs to generate the response based on dialogue context and the selected knowledge snippet. In our model, bi-directional attention is enabled within the knowledge and context parts to obtain better natural language understanding. As for the auto-regressive response generation, uni-directional attention is employed. The training objective is to minimize the negative log-likelihood (NLL) loss:
\begin{equation}
\mathcal{L}_{\rm generation} = -\log{p_{\rm generation}} (r |D_1,\cdots,D_t, k)
\end{equation}
where $r$ denotes the target response. During training, the ground-truth knowledge snippet is used for response generation. During inference, the knowledge snippet selected with Equation \eqref{eq:obj_sel} is used for response generation.

\section{Experiments}

\subsection{Multi-domain Dialogue State Tracking}

\begin{table*}[ht]
	\centering
	\includegraphics[width=\textwidth]{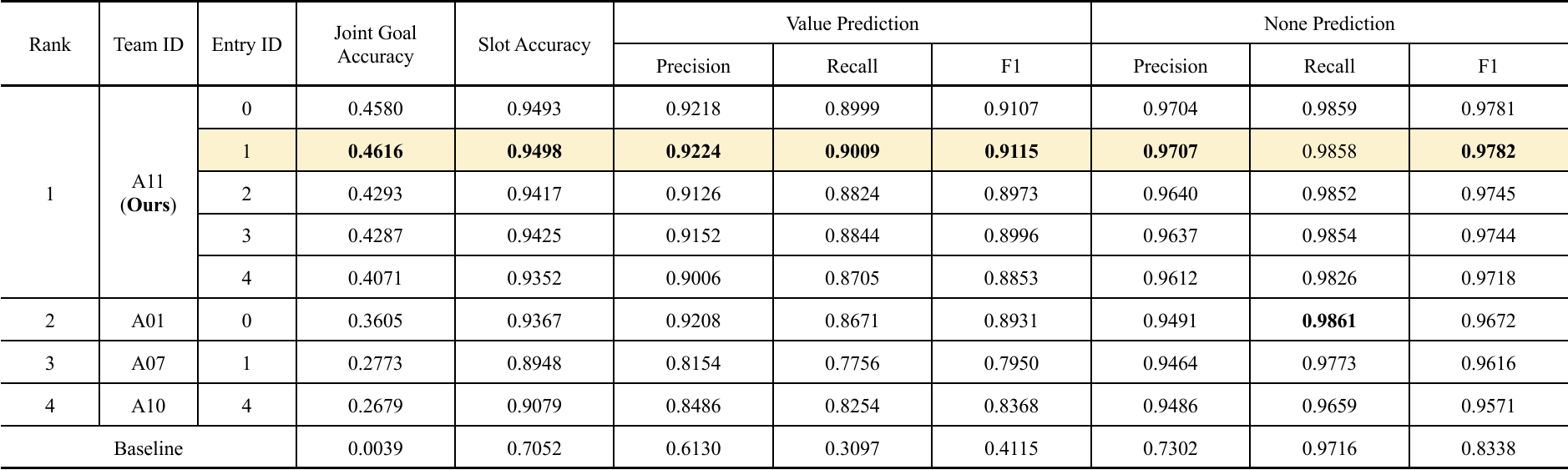}
	\caption{Final evaluation results on the test set of task 1, with the highest value written in bold.}
	\label{tab:dst_res}
\end{table*}

\subsubsection{Settings}

The dataset of DSTC10 Track2 includes spoken human-human dialogues about touristic information for San Francisco. In task 1, the validation set and test set share the same schema and database, where the validation set contains 107 dialogues, and the test set contains 783 dialogues. To tackle the lack of training data, we employ ontology based augmentation with MultiWOZ 2.3~\citep{han2020multiwoz}, pattern based augmentation with MultiWOZ 2.3 and validation set as well. Furthermore, we apply the SCS on the augmented train set to obtain noisy and spoken-like dialogues. Finally, the train set contains 376.9 thousand dialogues and 5.2 million utterances in total. In task 1, we use the train set to fine-tune the 1.6 billion parameters pre-trained dialogue generation model PLATO-2~\citep{bao2020plato}. Particularly, PLATO-2 is fine-tuned with a batch size of 8192 tokens and a learning rate of $1e-5$. All the models are trained on 8 Nvidia Telsa A100 40G GPU cards.

\subsubsection{Experimental Results}

For task 1, the details of the 5 entries we submitted are listed as follows:
\begin{itemize}
    \item \textbf{Entry 0}. The PLATO-2 is fine-tuned with a large amount of augmented training data. In addition to the generation model, the matching model based on BERT-Large~\citep{devlin2019bert} is also trained and utilized for the ensemble. The final result is an ensemble version of the multiple results by slot value voting with identical weight.~\footnotemark[3]
    \item \textbf{Entry 1}. The generation models and matching models are the same as entry 0. The final result is an ensemble version of the multiple results by slot value voting with the best weight estimated in the validation set.
    \item \textbf{Entry 2}. The generation models based on PLATO-2 are the same as entry 0. The final result is an ensemble version of the multiple results by slot value voting with identical weight.
    \item \textbf{Entry 3}. The generation models based on PLATO-2 are the same as entry 0. Moreover, a ranking model based on DeBERTa-XLarge~\citep{he2020deberta} is trained to rank multiple generated results. The final result is an ensemble version of the ranked results and the original results by slot value voting with identical weight.
    \item \textbf{Entry 4}. The single model based on PLATO-2 is fine-tuned to generate dialogue state.
\end{itemize}
\footnotetext[3]{In this ensemble, we carry out majority voting with these models to determine the value of each slot in each turn.}

In this task, the \textbf{joint goal accuracy (JGA)} is used as the final evaluation metric for the official ranking, where a sample is considered as correct if and only if all the slot values are correctly predicted. As shown in Table~\ref{tab:dst_res}, our system ranks first place with the best JGA. The results of all our entries, including the single generative model in entry 4, surpass the best entry of other teams. This benefits from the proposed data augmentation paradigm TOD-DA. A large amount of augmented training data produced by DE alleviate the problem of insufficient training data. SCS facilitates the effective learning of various oral expressions.

\begin{figure}
	\centering
	\includegraphics[width=\columnwidth]{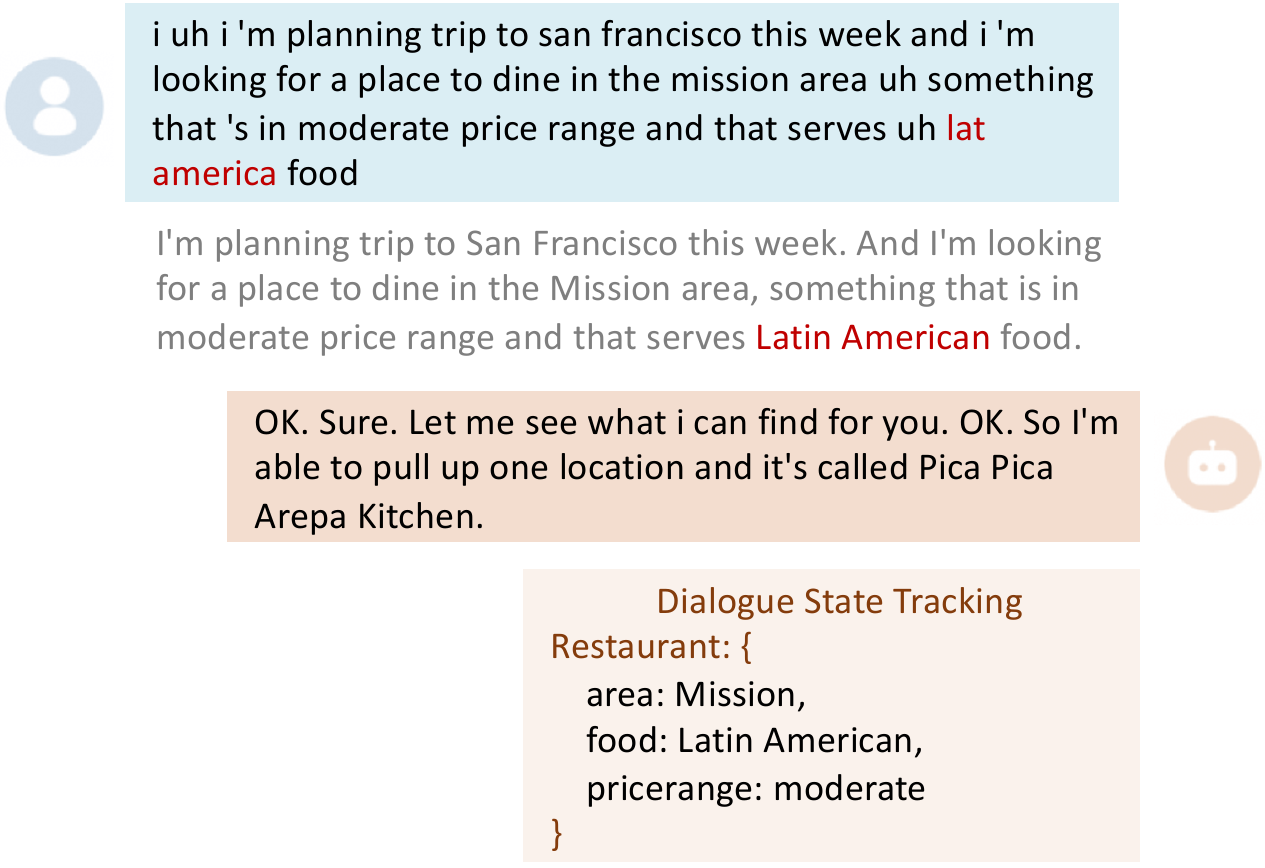}
	\caption{Case analysis on dialogue state tracking. Texts in \textcolor{gray}{gray} are corresponding manual transcriptions and texts in \textcolor{red}{red} indicate ASR errors.}
	\label{fig:dst_case}
\end{figure}

Figure~\ref{fig:dst_case} shows an example of spoken dialogue state tracking in the test set. It can be observed that the user utterance includes oral style expressions and ASR errors. Regardless, our model generates all dialogue states correctly for this example.

\subsection{Knowledge-grounded Dialogue Modeling}

\subsubsection{Settings}

To deal with the lack of train set, we apply the proposed ontology based augmentation on previous written conversations~\citep{kim2020domain}. Specifically, each ground-truth knowledge snippet and the corresponding entity in written conversations are replaced with new ones from the DSTC10 knowledge base. In subtask3, to imitate the verbal styles of looking up references during conversations, we extract some phrases from the validation set (e.g., \textit{``Let me check that for you."}) and randomly insert them into the augmented data. Furthermore, SCS is employed to get noisy and spoken-like dialogues. For our negatives enhanced knowledge selection, we use BM25 to calculate the semantic similarity.

The automatic evaluation of this task covers the following aspects in three cascaded subtasks:
\begin{itemize}
	\item \textbf{Subtask1 Knowledge-seeking Turn Detection.} Given the dialogue context, the model needs to decide whether to trigger external knowledge access or not. The automated metrics include Precision, Recall and F1.
	\item \textbf{Subtask2 Knowledge Selection.} For turns required external knowledge access, the model needs to select proper knowledge snippets from the knowledge base. The automated metrics include MRR@5~\citep{voorhees1999trec}, Recall@1 and Recall@5.
	\item \textbf{Subtask3 Knowledge-grounded Response Generation.} Given the dialogue context and selected knowledge snippets, the model needs to produce system responses. The automated metrics include BLEU-1/2/3/4~\citep{papineni2002bleu}, METEOR~\citep{denkowski2014meteor} and ROUGE-1/2/L~\citep{lin2004rouge}.
\end{itemize}

In the experiments, we fine-tune PLATO-2 for these three subtasks. Recently, the 11 billion parameter model PLATO-XL~\citep{bao2021plato} has shown its ability on task-oriented dialogues. We also utilize it to boost the response quality in knowledge-grounded generation.

\begin{table*}[ht!]
	\centering
	\includegraphics[width=\textwidth]{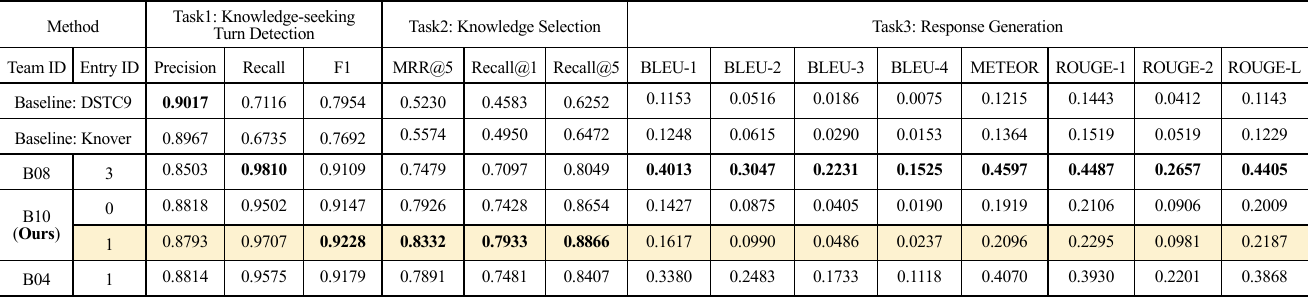}
	\caption{Automatic evaluation results on the test set of task 2, with the highest value written in bold.}
	\label{tab:auto_on_test}
\end{table*}

\subsubsection{Experimental Results}

The details of the two entries we submitted are described as follow:
\begin{itemize}
	\item \textbf{Entry 0.} All models of three subtasks are initialized with PLATO-2. 
	\item \textbf{Entry 1.} The model ensemble is carried out for both subtask1 and subtask2. PLATO-XL is used to initialize the generation model of subtask3.
\end{itemize}
In subtask3 of both entries, the response is generated using beam search, with a beam size of 5. To boost ensemble performance, we employ extra pre-training models including BERT-Large~\citep{devlin2019bert}, ALBERT-xxlarge~\citep{lan2019albert}, DeBERTa-Large and DeBERTa-XLarge~\citep{he2020deberta} in entry 1. For subtask1, the model ensemble is conducted through voting with identical weight. For subtask2, the other models re-rank the top-200 candidates from PLATO-2, and the final selection probability is the average result of all models.

The automatic evaluation results on the test set are summarized in Table~\ref{tab:auto_on_test}. There are two baselines: the official baseline from DSTC9~\citep{jin2021can} and Knover~\citep{he2021learning} from the DSTC9 track winner. With powerful pre-trained models and strengthened by the proposed TOD-DA, our models perform well on many automatic metrics. Our negatives enhanced strategy further boosts the performance of knowledge selection. Even the performance of our single model can be on par with that of the best ensemble entry from other teams.

\begin{table}
    \begin{center}
	\includegraphics[width=\columnwidth]{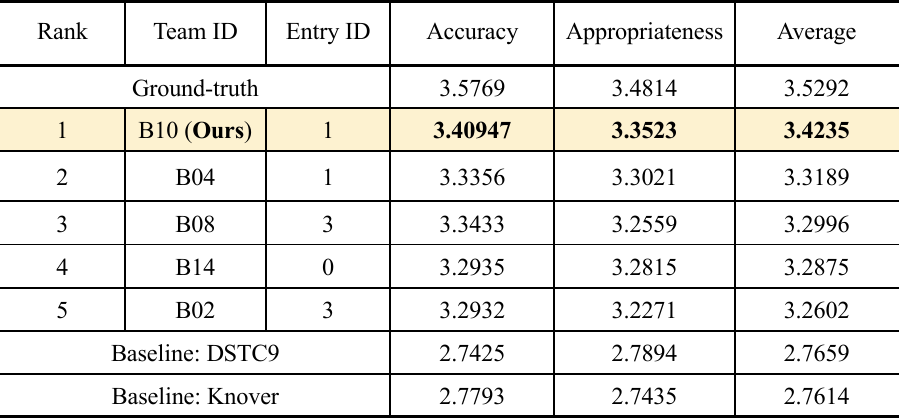}
	\end{center}
	\caption{Final human evaluation on the test set of task 2, with the highest value written in bold.}
	\label{tab:human_eval}
\end{table} 

In addition to the objective evaluation, human evaluation has been conducted for the final ranking. The quality of generated responses is evaluated by crowdsourcing workers under two metrics of \textbf{accuracy} and \textbf{appropriateness}. Accuracy measures how accurate the system output is based on the referenced knowledge snippets. Appropriateness evaluates how well the system output is naturally connected to the given conversation context. The evaluation score ranges from 1 to 5, with the higher, the better. The final evaluation results are shown in Table~\ref{tab:human_eval}. Our system ranks first place in the final results. The human ground-truth responses were also evaluated. The small gap between the ground-truth and our system suggests that the system can still provide high-quality and human-like services under the disturbance of ASR errors.

\begin{figure}
	\centering
	\includegraphics[width=\columnwidth]{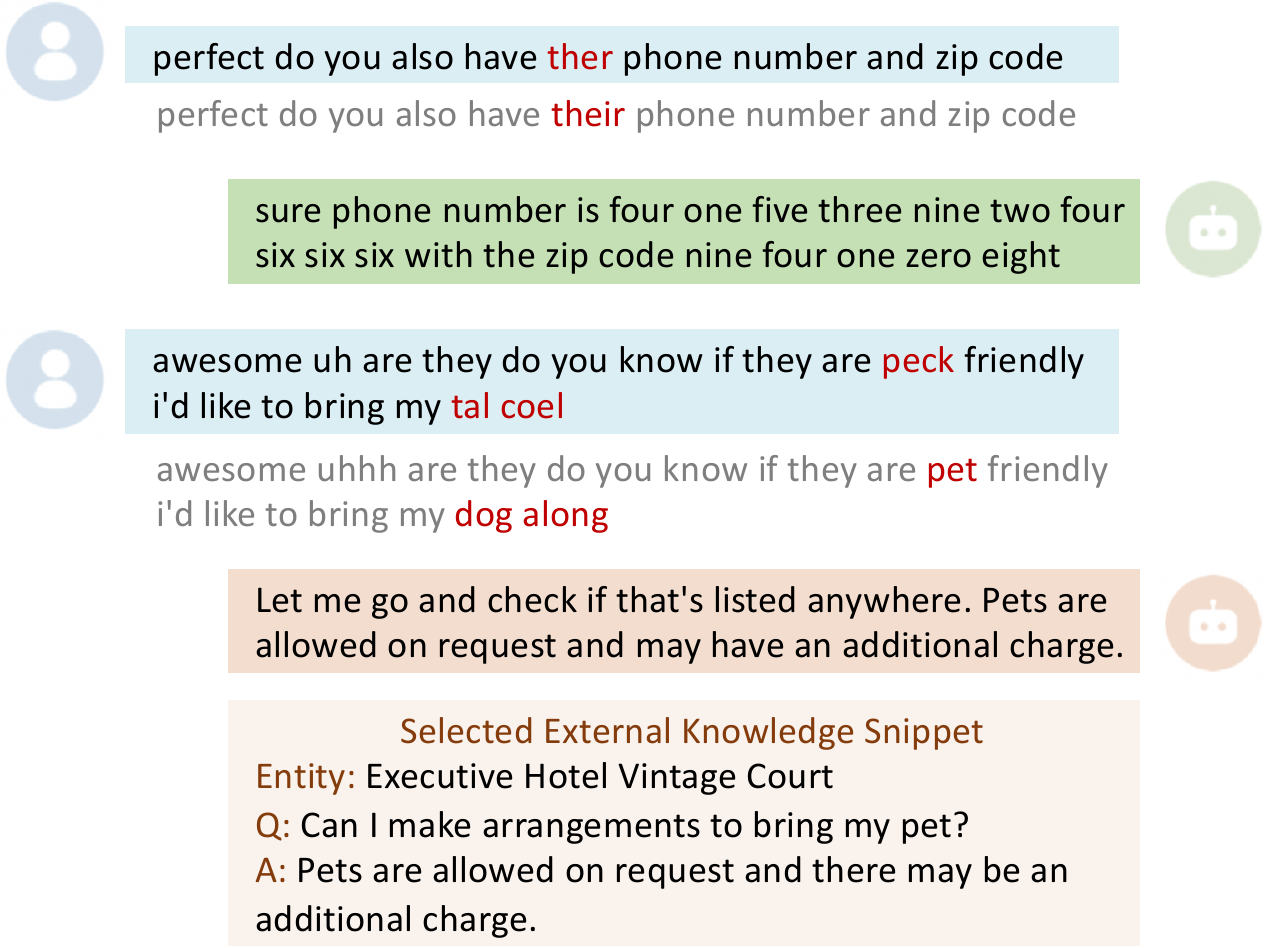}
	\caption{Case analysis on knowledge-grounded dialogue modeling. Texts in \textcolor{gray}{gray} are corresponding manual transcriptions and texts in \textcolor{red}{red} indicate ASR errors.}
	\label{fig:kgc_case}
\end{figure} 

To further analyze the performance of our system, one example is displayed in Figure~\ref{fig:kgc_case}. Even though the keywords in the dialogue context are eroded by ASR errors, the system can still locate the target knowledge snippet in the large knowledge base. By grounding on appropriate external knowledge, our system generates an accurate and human-like response.

\section{Related Work}

The related work will be discussed on data augmentation for task-oriented dialogue in two aspects: expanding dialogue datasets and simulating spoken conversations.

Expanding dialogue datasets is a way to alleviate the data insufficiency for task-oriented dialogue systems.
Some previous works leverage paraphrasing~\citep{wang2019task, iyyer2018adversarial, gao2020paraphrase} and backtranslation~\citep{sennrich2016improving, shleifer2019low} to rewrite the conversations without changing the slot values. Another trend to expand the datasets is randomly replacing the slot values without changing the original conversations, like SC-EDA in LAUG~\citep{liu-etal-2021-robustness}.
These works are confined to the fixed dialogue structure of the original datasets, resulting in the lack of diversity and flexibility in the expanded conversations.
In comparison, we propose Dialogue Enrichment (DE), a flexible data augmentation that gets exempt from restraints of the original dialogue structure. DE is able to produce a large amount of high-quality, fluent, and diverse task-oriented conversations with annotations.

Most of the publicly available task-oriented dialogue datasets are written conversations, which are insufficient to reflect actual human behaviors in practical spoken dialogue systems. To bridge the gap of the written and spoken conversations, many works attempt to imitate speech recognition errors and spoken conversation characteristics. To inject recognition noises, confusion-matrix-based ASR error simulators~\citep{wang2020data, gopalakrishnan2020neural} and TTS-ASR pipelines~\citep{liu-etal-2021-robustness} are applied to the original data. To imitate the disfluencies during speaking, some works~\citep{zayats2016disfluency, Wang_Che_Liu_Qin_Liu_Wang_2020} interrupt the original dialogues by inserting or deleting words and phrases. These works concentrate on word or sentence level simulations. In this paper, our proposed Spoken Conversation Simulator (SCS) imitates spoken conversations and ASR errors in more diverse granularities. The fine-grained phoneme-level simulator of SCS is utilized to ease errors caused by similar pronunciation during speech recognition.

\section{Conclusion}

In this paper, we propose Task-oriented Dialogue Data Augmentation (TOD-DA), a novel model-agnostic data augmentation paradigm for task-oriented dialogue on spoken conversations. Consisted of two modules: Dialogue Enrichment (DE) and Spoken Conversation Simulator (SCS), TOD-DA alleviates the problems of insufficient training data and the gap between spoken and written modalities in task-oriented dialogue. 
Our approach ranked first in both tasks of DSTC10 Track2, demonstrating the superiority and effectiveness of our proposed TOD-DA in boosting the robustness of task-oriented dialogue modeling on spoken conversations.

\section*{Acknowledgements}
We would like to thank the anonymous reviewers for their constructive suggestions; Chong Liu and Yipin Zhang for the helpful discussions. This work was supported by the Natural Key Research and Development Project of China (No. 2018AAA0101900).

\bibliography{bibtex}

\begin{thebibliography}{32}
\providecommand{\natexlab}[1]{#1}

\bibitem[{Baevski et~al.(2020)Baevski, Zhou, Mohamed, and
  Auli}]{NEURIPS2020_92d1e1eb}
Baevski, A.; Zhou, Y.; Mohamed, A.; and Auli, M. 2020.
\newblock wav2vec 2.0: A Framework for Self-Supervised Learning of Speech
  Representations.
\newblock In Larochelle, H.; Ranzato, M.; Hadsell, R.; Balcan, M.~F.; and Lin,
  H., eds., \emph{Advances in Neural Information Processing Systems},
  volume~33, 12449--12460. Curran Associates, Inc.

\bibitem[{Bao et~al.(2020)Bao, He, Wang, Wu, Wang, Wu, Guo, Liu, and
  Xu}]{bao2020plato}
Bao, S.; He, H.; Wang, F.; Wu, H.; Wang, H.; Wu, W.; Guo, Z.; Liu, Z.; and Xu,
  X. 2020.
\newblock Plato-2: Towards building an open-domain chatbot via curriculum
  learning.
\newblock \emph{arXiv preprint arXiv:2006.16779}.

\bibitem[{Bao et~al.(2021)Bao, He, Wang, Wu, Wang, Wu, Wu, Guo, Lu, Huang
  et~al.}]{bao2021plato}
Bao, S.; He, H.; Wang, F.; Wu, H.; Wang, H.; Wu, W.; Wu, Z.; Guo, Z.; Lu, H.;
  Huang, X.; et~al. 2021.
\newblock Plato-xl: Exploring the large-scale pre-training of dialogue
  generation.
\newblock \emph{arXiv preprint arXiv:2109.09519}.

\bibitem[{Budzianowski et~al.(2018)Budzianowski, Wen, Tseng, Casanueva, Ultes,
  Ramadan, and Ga{\v{s}}i{\'c}}]{budzianowski-etal-2018-multiwoz}
Budzianowski, P.; Wen, T.-H.; Tseng, B.-H.; Casanueva, I.; Ultes, S.; Ramadan,
  O.; and Ga{\v{s}}i{\'c}, M. 2018.
\newblock {M}ulti{WOZ} - A Large-Scale Multi-Domain {W}izard-of-{O}z Dataset
  for Task-Oriented Dialogue Modelling.
\newblock In \emph{Proceedings of the 2018 Conference on Empirical Methods in
  Natural Language Processing}, 5016--5026. Brussels, Belgium: Association for
  Computational Linguistics.

\bibitem[{Denkowski and Lavie(2014)}]{denkowski2014meteor}
Denkowski, M.; and Lavie, A. 2014.
\newblock Meteor universal: Language specific translation evaluation for any
  target language.
\newblock In \emph{Proceedings of the ninth workshop on statistical machine
  translation}, 376--380.

\bibitem[{Devlin et~al.(2019)Devlin, Chang, Lee, and
  Toutanova}]{devlin2019bert}
Devlin, J.; Chang, M.-W.; Lee, K.; and Toutanova, K. 2019.
\newblock BERT: Pre-training of Deep Bidirectional Transformers for Language
  Understanding.
\newblock In \emph{Proceedings of the 2019 Conference of the North American
  Chapter of the Association for Computational Linguistics: Human Language
  Technologies, Volume 1 (Long and Short Papers)}, 4171--4186.

\bibitem[{El~Asri et~al.(2017)El~Asri, Schulz, Sharma, Zumer, Harris, Fine,
  Mehrotra, and Suleman}]{el-asri-etal-2017-frames}
El~Asri, L.; Schulz, H.; Sharma, S.; Zumer, J.; Harris, J.; Fine, E.; Mehrotra,
  R.; and Suleman, K. 2017.
\newblock {F}rames: a corpus for adding memory to goal-oriented dialogue
  systems.
\newblock In \emph{Proceedings of the 18th Annual {SIG}dial Meeting on
  Discourse and Dialogue}, 207--219. Saarbr{\"u}cken, Germany: Association for
  Computational Linguistics.

\bibitem[{Gao et~al.(2020)Gao, Zhang, Ou, and Yu}]{gao2020paraphrase}
Gao, S.; Zhang, Y.; Ou, Z.; and Yu, Z. 2020.
\newblock Paraphrase Augmented Task-Oriented Dialog Generation.
\newblock In \emph{Proceedings of the 58th Annual Meeting of the Association
  for Computational Linguistics}, 639--649.

\bibitem[{Gopalakrishnan et~al.(2020)Gopalakrishnan, Hedayatnia, Wang, Liu, and
  Hakkani-Tur}]{gopalakrishnan2020neural}
Gopalakrishnan, K.; Hedayatnia, B.; Wang, L.; Liu, Y.; and Hakkani-Tur, D.
  2020.
\newblock Are neural open-domain dialog systems robust to speech recognition
  errors in the dialog history? an empirical study.
\newblock \emph{arXiv preprint arXiv:2008.07683}.

\bibitem[{Han et~al.(2020)Han, Liu, Takanobu, Lian, Huang, Wan, Peng, and
  Huang}]{han2020multiwoz}
Han, T.; Liu, X.; Takanobu, R.; Lian, Y.; Huang, C.; Wan, D.; Peng, W.; and
  Huang, M. 2020.
\newblock MultiWOZ 2.3: A multi-domain task-oriented dialogue dataset enhanced
  with annotation corrections and co-reference annotation.
\newblock \emph{arXiv preprint arXiv:2010.05594}.

\bibitem[{He et~al.(2021)He, Lu, Bao, Wang, Wu, Niu, and Wang}]{he2021learning}
He, H.; Lu, H.; Bao, S.; Wang, F.; Wu, H.; Niu, Z.; and Wang, H. 2021.
\newblock Learning to select external knowledge with multi-scale negative
  sampling.
\newblock \emph{arXiv preprint arXiv:2102.02096}.

\bibitem[{He et~al.(2020)He, Liu, Gao, and Chen}]{he2020deberta}
He, P.; Liu, X.; Gao, J.; and Chen, W. 2020.
\newblock Deberta: Decoding-enhanced bert with disentangled attention.
\newblock In \emph{International Conference on Learning Representations}.

\bibitem[{Iyyer et~al.(2018)Iyyer, Wieting, Gimpel, and
  Zettlemoyer}]{iyyer2018adversarial}
Iyyer, M.; Wieting, J.; Gimpel, K.; and Zettlemoyer, L. 2018.
\newblock Adversarial Example Generation with Syntactically Controlled
  Paraphrase Networks.
\newblock In \emph{Proceedings of the 2018 Conference of the North American
  Chapter of the Association for Computational Linguistics: Human Language
  Technologies, Volume 1 (Long Papers)}, 1875--1885.

\bibitem[{Jin, Kim, and Hakkani-Tur(2021)}]{jin2021can}
Jin, D.; Kim, S.; and Hakkani-Tur, D. 2021.
\newblock Can I Be of Further Assistance? Using Unstructured Knowledge Access
  to Improve Task-oriented Conversational Modeling.
\newblock \emph{arXiv preprint arXiv:2106.09174}.

\bibitem[{Kim et~al.(2020{\natexlab{a}})Kim, Eric, Gopalakrishnan, Hedayatnia,
  Liu, and Hakkani-Tur}]{kim-etal-2020-beyond}
Kim, S.; Eric, M.; Gopalakrishnan, K.; Hedayatnia, B.; Liu, Y.; and
  Hakkani-Tur, D. 2020{\natexlab{a}}.
\newblock Beyond Domain {API}s: Task-oriented Conversational Modeling with
  Unstructured Knowledge Access.
\newblock In \emph{Proceedings of the 21th Annual Meeting of the Special
  Interest Group on Discourse and Dialogue}, 278--289. 1st virtual meeting:
  Association for Computational Linguistics.

\bibitem[{Kim et~al.(2020{\natexlab{b}})Kim, Eric, Gopalakrishnan, Hedayatnia,
  Liu, and Hakkani-Tur}]{kim2020domain}
Kim, S.; Eric, M.; Gopalakrishnan, K.; Hedayatnia, B.; Liu, Y.; and
  Hakkani-Tur, D. 2020{\natexlab{b}}.
\newblock Beyond Domain APIs: Task-oriented Conversational Modeling with
  Unstructured Knowledge Access.
\newblock \emph{arXiv preprint arXiv:2006.03533}.

\bibitem[{Kim et~al.(2021)Kim, Liu, Jin, Papangelis, Gopalakrishnan,
  Hedayatnia, and Hakkani-Tur}]{kim2021how}
Kim, S.; Liu, Y.; Jin, D.; Papangelis, A.; Gopalakrishnan, K.; Hedayatnia, B.;
  and Hakkani-Tur, D. 2021.
\newblock "How robust r u?": Evaluating Task-Oriented Dialogue Systems on
  Spoken Conversations.
\newblock arXiv:2109.13489.

\bibitem[{Lan et~al.(2019)Lan, Chen, Goodman, Gimpel, Sharma, and
  Soricut}]{lan2019albert}
Lan, Z.; Chen, M.; Goodman, S.; Gimpel, K.; Sharma, P.; and Soricut, R. 2019.
\newblock ALBERT: A Lite BERT for Self-supervised Learning of Language
  Representations.
\newblock In \emph{International Conference on Learning Representations}.

\bibitem[{Lin(2004)}]{lin2004rouge}
Lin, C.-Y. 2004.
\newblock Rouge: A package for automatic evaluation of summaries.
\newblock In \emph{Text summarization branches out}, 74--81.

\bibitem[{Liu et~al.(2021)Liu, Takanobu, Wen, Wan, Li, Nie, Li, Peng, and
  Huang}]{liu-etal-2021-robustness}
Liu, J.; Takanobu, R.; Wen, J.; Wan, D.; Li, H.; Nie, W.; Li, C.; Peng, W.; and
  Huang, M. 2021.
\newblock Robustness Testing of Language Understanding in Task-Oriented Dialog.
\newblock In \emph{Proceedings of the 59th Annual Meeting of the Association
  for Computational Linguistics and the 11th International Joint Conference on
  Natural Language Processing (Volume 1: Long Papers)}, 2467--2480. Online:
  Association for Computational Linguistics.

\bibitem[{Panayotov et~al.(2015)Panayotov, Chen, Povey, and
  Khudanpur}]{panayotov-chen-2015-librispeech}
Panayotov, V.; Chen, G.; Povey, D.; and Khudanpur, S. 2015.
\newblock Librispeech: An ASR corpus based on public domain audio books.
\newblock In \emph{2015 IEEE International Conference on Acoustics, Speech and
  Signal Processing (ICASSP)}, 5206--5210.

\bibitem[{Papineni et~al.(2002)Papineni, Roukos, Ward, and
  Zhu}]{papineni2002bleu}
Papineni, K.; Roukos, S.; Ward, T.; and Zhu, W.-J. 2002.
\newblock Bleu: a method for automatic evaluation of machine translation.
\newblock In \emph{Proceedings of the 40th annual meeting of the Association
  for Computational Linguistics}, 311--318.

\bibitem[{Sennrich, Haddow, and Birch(2016)}]{sennrich2016improving}
Sennrich, R.; Haddow, B.; and Birch, A. 2016.
\newblock Improving Neural Machine Translation Models with Monolingual Data.
\newblock In \emph{Proceedings of the 54th Annual Meeting of the Association
  for Computational Linguistics (Volume 1: Long Papers)}, 86--96.

\bibitem[{Shleifer(2019)}]{shleifer2019low}
Shleifer, S. 2019.
\newblock Low resource text classification with ulmfit and backtranslation.
\newblock \emph{arXiv preprint arXiv:1903.09244}.

\bibitem[{Vaswani et~al.(2017)Vaswani, Shazeer, Parmar, Uszkoreit, Jones,
  Gomez, Kaiser, and Polosukhin}]{vaswani2017attention}
Vaswani, A.; Shazeer, N.; Parmar, N.; Uszkoreit, J.; Jones, L.; Gomez, A.~N.;
  Kaiser, {\L}.; and Polosukhin, I. 2017.
\newblock Attention is all you need.
\newblock In \emph{Advances in neural information processing systems},
  5998--6008.

\bibitem[{Voorhees(1999)}]{voorhees1999trec}
Voorhees, E.~M. 1999.
\newblock The TREC-8 Question Answering Track Report.
\newblock In \emph{Proceedings of the 8th Text REtrieval Conference (TREC-8),
  1999}.

\bibitem[{Wang et~al.(2020{\natexlab{a}})Wang, Fazel-Zarandi, Tiwari,
  Matsoukas, and Polymenakos}]{wang2020data}
Wang, L.; Fazel-Zarandi, M.; Tiwari, A.; Matsoukas, S.; and Polymenakos, L.
  2020{\natexlab{a}}.
\newblock Data Augmentation for Training Dialog Models Robust to Speech
  Recognition Errors.
\newblock In \emph{Proceedings of the 2nd Workshop on Natural Language
  Processing for Conversational AI}, 63--70.

\bibitem[{Wang et~al.(2020{\natexlab{b}})Wang, Che, Liu, Qin, Liu, and
  Wang}]{Wang_Che_Liu_Qin_Liu_Wang_2020}
Wang, S.; Che, W.; Liu, Q.; Qin, P.; Liu, T.; and Wang, W.~Y.
  2020{\natexlab{b}}.
\newblock Multi-Task Self-Supervised Learning for Disfluency Detection.
\newblock \emph{Proceedings of the AAAI Conference on Artificial Intelligence},
  34(05): 9193--9200.

\bibitem[{Wang et~al.(2019)Wang, Gupta, Chang, and Baldridge}]{wang2019task}
Wang, S.; Gupta, R.; Chang, N.; and Baldridge, J. 2019.
\newblock A task in a suit and a tie: paraphrase generation with semantic
  augmentation.
\newblock In \emph{Proceedings of the AAAI Conference on Artificial
  Intelligence}, volume~33, 7176--7183.

\bibitem[{Wei and Zou(2019)}]{wei-zou-2019-eda}
Wei, J.; and Zou, K. 2019.
\newblock {EDA}: Easy Data Augmentation Techniques for Boosting Performance on
  Text Classification Tasks.
\newblock In \emph{Proceedings of the 2019 Conference on Empirical Methods in
  Natural Language Processing and the 9th International Joint Conference on
  Natural Language Processing (EMNLP-IJCNLP)}, 6382--6388. Hong Kong, China:
  Association for Computational Linguistics.

\bibitem[{Wen et~al.(2017)Wen, Vandyke, Mrk{\v{s}}i{\'c}, Ga{\v{s}}i{\'c},
  Rojas-Barahona, Su, Ultes, and Young}]{wen-etal-2017-network}
Wen, T.-H.; Vandyke, D.; Mrk{\v{s}}i{\'c}, N.; Ga{\v{s}}i{\'c}, M.;
  Rojas-Barahona, L.~M.; Su, P.-H.; Ultes, S.; and Young, S. 2017.
\newblock A Network-based End-to-End Trainable Task-oriented Dialogue System.
\newblock In \emph{Proceedings of the 15th Conference of the {E}uropean Chapter
  of the Association for Computational Linguistics: Volume 1, Long Papers},
  438--449. Valencia, Spain: Association for Computational Linguistics.

\bibitem[{Zayats, Ostendorf, and Hajishirzi(2016)}]{zayats2016disfluency}
Zayats, V.; Ostendorf, M.; and Hajishirzi, H. 2016.
\newblock Disfluency detection using a bidirectional lstm.
\newblock \emph{arXiv preprint arXiv:1604.03209}.

\end{thebibliography}

\end{document}